\documentclass[lettersize,journal]{IEEEtran}
\usepackage{times}
\usepackage{helvet}
\usepackage{courier}
\usepackage{amsmath,amsfonts}
\usepackage{amsthm}
\usepackage{algorithmic}
\usepackage{algorithm}  
\usepackage{array}
\usepackage{textcomp}
\usepackage{url}
\usepackage{verbatim}
\usepackage{graphicx}
\usepackage{enumerate}
\usepackage{subfigure}
\usepackage{booktabs}
\usepackage[table]{xcolor} 
\usepackage{bbding}
\usepackage{multirow}
\usepackage{multicol}
\usepackage[caption=false,font=footnotesize,labelfont=rm,textfont=rm]{subfig}
\usepackage{cite}

\usepackage{xcolor}

\usepackage{balance}
\begin{document}

\title{DA-Flow: Dual Attention Normalizing Flow for Skeleton-based Video Anomaly Detection}
 \author{Ruituo Wu, Yang Chen, Jian Xiao, Bing Li, \emph{Member} \emph{IEEE}, Jicong Fan, \emph{Senior Member} \emph{IEEE}, \\ Frédéric Dufaux, \emph{Fellow} \emph{IEEE}, Ce Zhu, \emph{Fellow} \emph{IEEE}, and Yipeng Liu, \emph{Senior Member} \emph{IEEE}

\thanks{This work was supported in part by the National Natural Science Foundation of China (NSFC) under Grant 62171088 and 62376236. Yipeng Liu is the corresponding author.}

\thanks{Ruituo Wu, Yang Chen, Bing Li, Ce Zhu, and Yipeng Liu are with the School of Information and Communication Engineering, University of Electronic Science and Technology of China (UESTC), Chengdu, 611731, China. E-mail: yipengliu@uestc.edu.cn.}
\thanks{Jian Xiao is with the School of Computer and Electronic Information /Artificial Intelligence,
Nanjing Normal University, Nanjing, 210023, China}
\thanks{Jicong Fan is with the Chinese University of Hong Kong, Shenzhen, 518172, China}
\thanks{F. Dufaux is with the Laboratoire des Signaux et Systemes, Universite Paris-Saclay, CNRS, CentraleSupelec, 91192, Gif-sur-Yvette, France}}
\markboth{Journal of \LaTeX\ Class Files,~Vol.~18, No.~9, September~2020}%
{How to Use the IEEEtran \LaTeX \ Templates}
\maketitle
\begin{abstract}
Cooperation between temporal convolutional networks (TCN) and graph convolutional networks (GCN) as a processing module has shown promising results in skeleton-based video anomaly detection (SVAD).  However, to maintain a lightweight model with low computational and storage complexity, shallow GCN and TCN blocks are constrained by small receptive fields and a lack of cross-dimension interaction capture.  To tackle this limitation, we propose a lightweight module called the Dual Attention Module (DAM) for capturing cross-dimension interaction relationships in spatio-temporal skeletal data. It employs the frame attention mechanism to identify the most significant frames and the skeleton attention mechanism to capture broader relationships across fixed partitions with minimal parameters and flops. Furthermore, the proposed Dual Attention Normalizing Flow (DA-Flow)  integrates the DAM as a post-processing unit after GCN within the normalizing flow framework. Simulations show that the proposed model is robust against noise and negative samples. Experimental results show that DA-Flow reaches competitive or better performance than the existing state-of-the-art (SOTA) methods in terms of the micro AUC metric with the fewest number of parameters. Moreover, we found that even without training, simply using random projection without dimensionality reduction on skeleton data enables substantial anomaly detection capabilities.
\end{abstract}
\section{Introduction}
\label{sec:intro}
Video Anomaly Detection (VAD) identifies abnormal events within video streams, such as instances of violence or emergencies, which plays an increasingly important role in video surveillance~\cite{37pred}.  Given the vast spectrum of anomalies present in real-life situations, it is infeasible to collect samples of every conceivable anomaly for supervised learning purposes. Because of the accessibility and ease of acquiring normal video data along with its annotations, there has been a surge of interest in detecting video anomalies merely trained on accessible normal samples. 
If an algorithm struggles with imbalanced data, it might wrongly flag rare normal events as abnormal, leading to false positives \cite{cao2024twostream}.

Existing VAD methods can be broadly classified based on the data used: raw image/video data or pose/skeleton data. The former directly uses RGB images or video streams~\cite{36convae,ssmtl,normalgraph,SSMTL++} as the input of the detection model. While being straightforward and effective, this approach requires large models to filter out irrelevant features and identify anomalies, leading to high computational and storage costs, and exacerbating imbalanced minority anomaly problems~\cite{zhang2021smaller,tan2019imbalance}.

\begin{figure}[!t]
   \centering
   \includegraphics[scale=0.45]{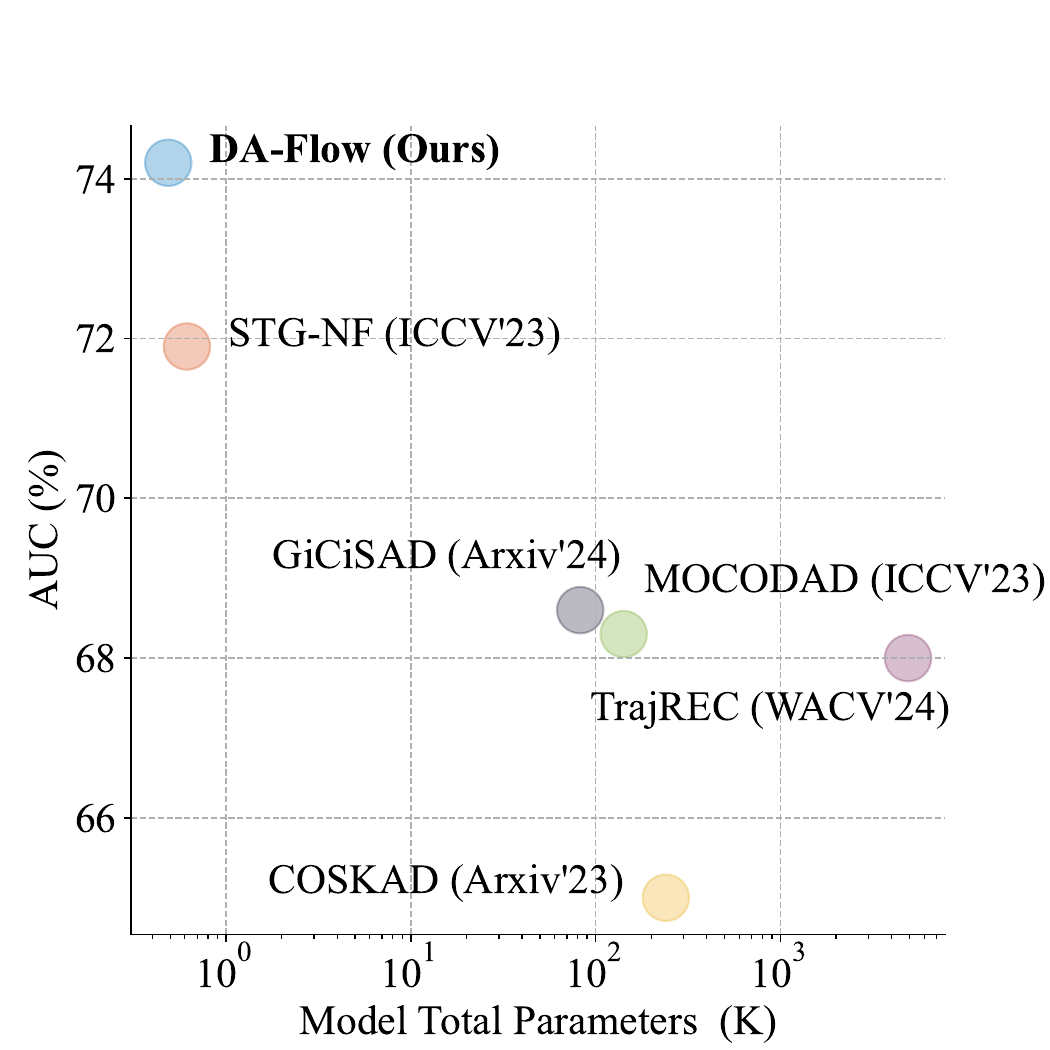}
    \caption{The comparison between DA-Flow with State-of-the-Art Skeleton-based Video Anomaly Detection methods (STG-NF \cite{stgnf}, MOCODAD \cite{mocodad}, COSKAD \cite{COSKAD},TrajREC \cite{stergiou2024traj}) and GiCiSAD \cite{karami2024gici} on UBnormal dataset. Our method surpasses these methods on the AUC metric with the fewest number of parameters.} 
    \label{ub}

\end{figure}

Observing the drawbacks of image/video-based methods, a recent trend of VAD builds on top of pose and skeleton data~\cite{mpedrnn,multi,posecvae,bipoco,stgcaelstm,gepc,stgnf}. In contrast with traditional image/video-based methods that keep all raw data intact, skeleton data offer an informative, compact, and well-structured representation that is beneficial to the VAD task, characterized by intuitive low-dimensional signals.  The advances of pose/skeleton extraction models such as Poseflow~\cite{33poseflow} and Alphapose~\cite{34alphapose} are driving increasing feasibility and effectiveness of skeleton-based video anomaly detection (SVAD), which facilitates sidestepping the complexities of intricate raw image/video data and democratizing VAD task. 
 Extracting information specific to human actions cannot only provide a privacy-protecting solution but can also help filter out the background-related noise in the videos and aid the model to focus on key information for detecting abnormal events related to human behavior. Its reduced computational demand makes it ideal for resource-limited environments and situations requiring quick responses such as monitoring in homes or nursing facilities where privacy is paramount. By focusing on skeletal representations rather than detailed visual images, skeleton-based approaches inherently reduce the amount of sensitive information processed \cite{stgnf}. In addition, models with a smaller number of parameters have better generalization ability \cite{zhang2021smaller} and the capacity to better cope with unbalanced samples \cite{tan2019imbalance}.

The majority of SVAD methods employ graph convolutional networks (GCN)~\cite{normalgraph,zeng2021hierarchical,wu2021gcnmix,35gcn} to extract local features from skeletons and $1 \times 1$ temporal convolutional networks (TCN) \cite{stgnf,normalgraph,zeng2021hierarchical} or a multi-scale architectural approach \cite{chen2023multiscale}.  However, GCN and TCN face two challenges in SVAD: Firstly, GCN's reliance on heuristic partitioning restricts its ability to capture interactions between distinct body parts, such as the head and hand which leads to insufficient connectivity when these parts are not grouped within the same partition. Additionally, TCN's post-convolution receptive field remains limited due to its $1 \times 1$ convolution operation. Secondly, both GCN and TCN inadequately capture the intricate interactions across the three dimensions inherent in spatial-temporal skeleton data, which is structured into channels, skeletal joints, and frames: including the relationships between channels and skeletal joints, and between channels and frames. This limitation is particularly evident in applications like fall detection, where variations along the vertical axis are more pronounced. It is crucial to address these limitations to enhance the understanding of complex spatial-temporal relationships in skeleton data.

Moreover, real-time capability holds paramount significance in the context of SVAD, given that anomalies typically necessitate prompt intervention. The computational overhead of multi-scale structures designed by stacking convolutional layers to increase the receptive field and information capture capability is unacceptable. Hence, it becomes imperative to employ a lightweight solution characterized by minimal parameters and computational overhead. 

To resolve the above-outlined challenges, inspired by the effectiveness of non-local attention \cite{16} in video detection and the success of attention mechanism across channels \cite{17}, 
we introduce a lightweight attention mechanism called the dual attention module (DAM). It is specifically designed for spatio-temporal skeletal data in SVAD. This mechanism is crafted to be efficient, enhancing the model's ability to focus on critical features within skeletal data for improved performance. This module incorporates a dual-branch global attention mechanism for improved post-GCN processing:

\begin{enumerate}[1.]
\item The first branch, known as \textbf{\textit{Skeleton Attention}}, identifies `\textbf{\textit{Which skeletal joints are more crucial}}' through capturing channel-skeletal joint relationships. \textbf{\textit{Skeleton Attention}} capture the behaviors between different joints across GCN fixed partitions of fixed frames.
\end{enumerate}
\begin{enumerate}[2.]
\item The second branch, termed \textbf{\textit{Frame Attention}},  focuses on determining `\textbf{\textit{Which frames are more crucial}}' by attempting to capture channel-frame relationships. 
\end{enumerate}
These two attention mechanisms collaboratively capture spatio-temporal co-occurrence relationships. Each branch uses a lightweight and effective attention mechanism. First, permute the input data and apply max pooling to extract the `sharpest' features. Then, utilize 2-D convolution to generate an attention `map'. Finally, implement the attention mechanism by broadcast multiplying the input tensor with its corresponding attention `map'. We take the average of the \textbf{\textit{Skeleton Attention}} and the \textbf{\textit{Frame Attention}} as the output of DAM. To the best of our knowledge, DAM is the first attention mechanism designed specifically for lightweight SVAD solutions.

Furthermore, we integrate the DAM after the GCN phase to jointly capture global and local information, serving as a core transformation unit to Glow \cite{27glow} and compute the minimal log-likelihood across all individuals in a frame to detect anomalies. Extensive experiments on five video anomaly detection datasets: ShanghaiTech, HR-ShanghaiTech, UB-normal, HR-UBnormal, and UCSD Ped2 demonstrate that our model outperforms previous state-of-the-art methods with by far the fewest number of parameters in SVAD (as shown in Fig.~\ref{ub}). Additionally, our experiments with contaminated data validate the robustness of our model. Finally, we adopt a unique approach by not training our model. Instead, we test it immediately after a random setup. This approach goes beyond several image-based VAD results, demonstrating that normal and abnormal behavior in skeleton data exhibit different statistical characteristics, making them suitable for VAD. Based on this observation, we believe that the focus on SVAD in future work should not be solely on reconstruction and prediction, or a combination, but rather focusing on the statistical characteristics of normal samples.

The main contributions of this paper are as follows:
\begin{itemize}
 \item  We propose a lightweight and effective attention mechanism designed for SVAD named DAM to capture the spatio-temporal co-occurrence relationship by incorporating \textbf{\textit{Skeleton Attention}} and  \textbf{\textit{Frame Attention}}.
 \item We integrate the proposed DAM post-GCN within a Glow-like structure as a core transformation unit to jointly capture both global and local information,  and build a new effective method for skeleton-based video anomaly detection with by far the fewest number of parameters (only 0.488K).
  \item  Experiments show that our approach reaches superior results compared to the state-of-the-art methods with by far the fewest number of parameters in SVAD on five benchmarks.
  \item  We discover that even without training, simply employing random projection without dimensionality reduction on skeleton data enables us to achieve substantial anomaly detection capabilities. 
\end{itemize}

\section{Related Work}
\subsection{Skeleton-based Video Anomaly Detection }

SVAD aims to identify unusual human behaviors in video footage using skeletal data. By leveraging learning characteristics of human skeletal data, it discerns normal behavior patterns from skeletal information and flags deviations as anomalies. These approaches are predominantly categorized into four broad categories, i.e., reconstruction, prediction, their combinations, and modeling distribution.

\textit{Reconstruction-based methods} within the context of SVAD predominantly utilize Convolutional Autoencoders (CAE) to emulate typical video behaviors. These approaches identify anomalies via increased reconstruction errors and diverge from established normative patterns, thereby utilizing this discrepancy as a criterion for anomaly detection.  \cite{jiang2022GRU} introduced a framework incorporating a Gated Recurrent Unit (GRU) encoder-decoder network, adept at detecting and pinpointing anomalous pedestrian behaviors in videos captured at the grade crossing. Furthermore, \cite{fan2022GAN} devised an anomaly detection architecture comprising dual generator and discriminator pairs, where the generators were tasked with reconstructing normal video frames and their skeletal counterparts. The discriminators aimed to differentiate between the original and reconstructed entities for both video frames and skeletons. 

\textit{Prediction-based methods} focus on learning normal human behavior by forecasting future actions based on past observations. Specifically, these methods train a network to predict future skeletons from sequences representing normal human activities. During the evaluation, samples with significant prediction errors are classified as anomalies. \cite{normalgraph} introduced a GCN-based method, leveraging the spatial and temporal dimensions of human movements. \cite{zeng2021hierarchical} developed a more complex model that captures both high-level interactions among individuals and low-level postures of each person by utilizing a hierarchical GCN structure. \cite{huang2022transformer} presented a novel approach using transformers for encoding hierarchical graph embeddings which focused on both individual and inter-individual correlations.

\textit{Combination-based methods} merge reconstruction and prediction techniques to model normal human behaviors. \cite{morais2019mix} decomposed human skeletal data into global and local components and then modeled them as two interacting subprocesses thus offering a comprehensive view of normal human movements. \cite{li2022grumix} introduced a novel single-encoder-dual-decoder GRU architecture. One decoder focuses on reconstructing the input skeletons, while the other predicts future skeletons.  \cite{li2022comlstm} introduced a single-encoder-dual-decoder architecture based on a spatio-temporal Graph CAE (GCAE) with LSTM networks in hidden layers. 

\textit{Modeling distribution-based} methods focus on identifying outliers by analyzing the distribution or clustering of skeleton graphs. \cite{tani2022gaus} utilized features extracted from skeletons exhibiting normal behavior to model a multivariate Gaussian distribution. \cite{stgnf} utilizes normalizing flows to capture characteristics of typical skeletal movements and enable the identification of anomalies by evaluating adherence to a normal data distribution. \cite{chen2023multiscale}  features an encoder for feature extraction, a reconstruction decoder to refine the encoder's performance, and a clustering layer to determine anomaly scores.

\subsection{Normalizing Flow}

Normalizing flow represents a transformative approach in the domain of generative modeling by using invertible transformations to build complex probability distributions.

Initially, NICE \cite{28nice} introduced NF to deep learning with a simple architecture that partitions input dimensions and employs additive coupling layers for efficient, invertible modeling. Subsequently, NVP \cite{29NVP} enhanced flexibility by integrating additive and multiplicative transformations within its coupling layers, capturing more intricate distributions. Further advancements came with Neural Spline Flows (NSF) \cite{30nsf}, which use spline-based transformations to model non-linear relationships more effectively, refining generative models' expressiveness. Once trained on `normal' data, NF models assign higher likelihoods to similar new samples and lower likelihoods to outliers, enabling effective anomaly detection.

\subsection{Convolutional Attention Mechanism}

The convolutional attention mechanism marks a significant advance in deep learning, particularly for enhancing CNNs' interpretability and performance. By focusing dynamically on salient features within an input, such mechanisms enable more effective learning of relevant patterns. Notable architectures that have emerged include CBAM \cite{woo2018cbam} (Convolutional Block Attention Module), SENet \cite{hu2018senet} (Squeeze-and-Excitation Networks), Coordinate Attention \cite{hou2021coordinate}, and Triplet Attention \cite{17}.

CBAM \cite{woo2018cbam} integrates attention mechanisms both spatially and channel-wise within CNNs, refining feature maps sequentially through focus on relevant channels followed by attention to important spatial regions. SENet \cite{hu2018senet} pioneered the concept of explicitly modeling interdependencies between channels, and employs a mechanism that adaptively recalibrates channel-wise feature responses. Unlike traditional approaches that aggregate global information indiscriminately, Triplet Attention \cite{17} extends the attention mechanism across three dimensions—channel, height, and width—simultaneously, offering a comprehensive approach to focus within the network. Coordinate Attention \cite{hou2021coordinate} introduces a novel perspective by incorporating positional information into the attention mechanism. Furthermore, \cite{song2022skeatt} applies the Coordinate Attention to skeleton-based action recognition. 
But to our knowledge, no work has considered lightweight attention mechanisms dedicated to SVAD solutions.
\begin{figure*}[!t]
   \centering
   \includegraphics[scale=0.95]{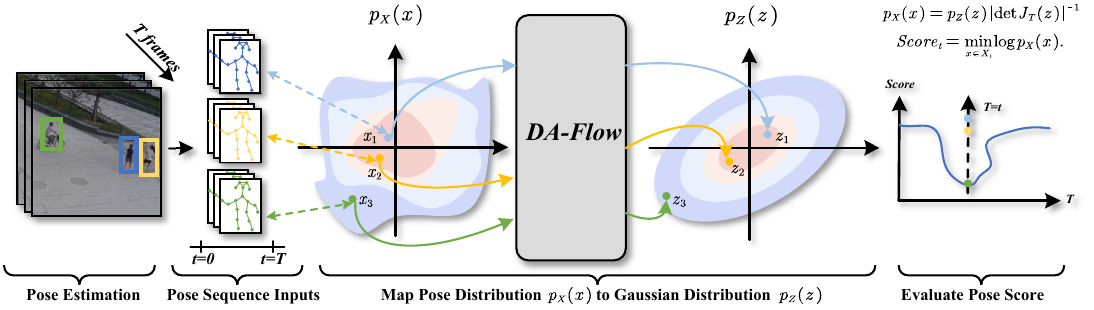}
   \caption{ Overview of the DA-flow Methodology. The process initiates with pose estimation and tracking of input video data. Each pose sequence is then individually processed by the DA-flow model. Training involves learning a bijective mapping from the data distribution \( p_X(x) \) (pose sequences) to a latent Gaussian prior \( p_Z(z) \), achieved by minimizing the negative log-likelihood of the data. This utilizes the invertibility of the architecture along with the change of variables formula. During inference, the probability of each pose sequence is assessed, and the frame score is determined by the sequence with the lowest log-likelihood score. }

   \label{overvieww}
\end{figure*}

\section{Methodology}
\subsection{Overview}
An overview of our method is shown in Fig.~\ref{overvieww}: 
Given a video sequence, we employ a standard human pose detector and tracker to extract poses, which are subsequently represented as spatial-temporal skeletal graphs. During training, DA-Flow learns a bijective mapping from data distribution \( p_X \) (pose sequences) to a latent Gaussian distribution \( p_Z \) by using the change of variables formula. For inference, we estimate the probability of each pose sequence by reverse calculation.

Assuming a video with \(T\) frames, we consider a specific frame at time \(T=t\) containing multiple individuals, denoted as \(X_t\). For each individual represented by a skeletal diagram \(x\) in \(X_t\), we compute the score by evaluating \(\log p_X(x)\). The overall score for the frame denoted as \(\operatorname{Score}_t\) is then defined as the minimum of these computed values across all individuals:
\begin{equation}
\operatorname{Score}_t = \min_{x \in X_t} \log p_X(x).
\end{equation}

\begin{figure*}[!tbp]
   \centering
   \includegraphics[scale=1]{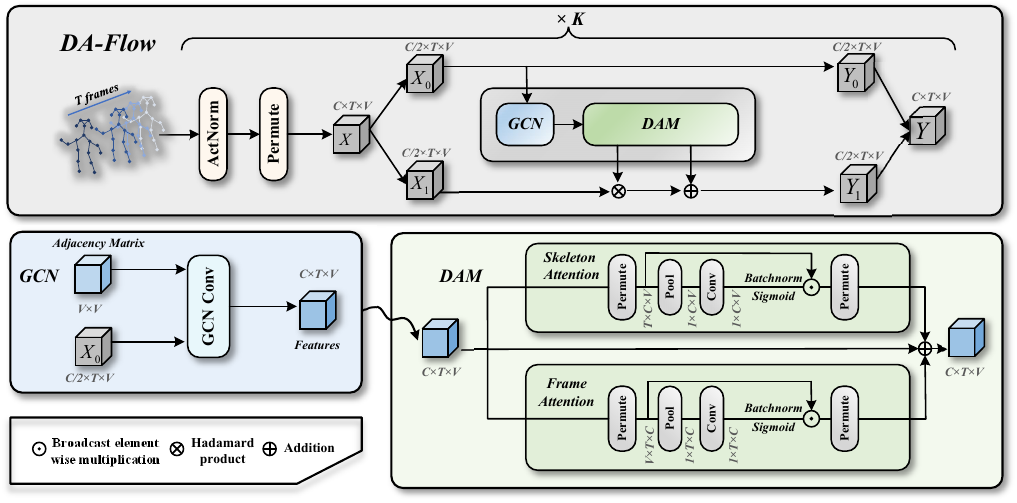}

    \caption{Schematic diagram of the DA Flow includes Actnorm, permutation, and affine coupling layers in each transformation unit. A transformation unit begins with a GCN that extracts local information from a heuristic partition of the human skeleton. The subsequent DAM consists of two main branches: the upper branch dedicated to extracting \textbf{\textit{Skeleton Attention}}, and the lower branch, which splits further to concurrently extract \textbf{\textit{Frame Attention}}.
    }

\label{block}
\end{figure*}

In DA-flow, each flow step integrates an Actnorm layer, a permutation layer, and an affine coupling layer within a framework combining GCN and DAM. Actnorm, functioning similarly to batch normalization \cite{27glow}, normalizes the activations. The permutation layer employs a \(1 \times 1\) convolution for reversible permutations. The affine coupling layer splits the input data channel-wise and helps in facilitating bijective transformations and Jacobian calculations. In our framework, GCN first extracts local features from skeletal data to capture specific details. Subsequently, DAM extends this by identifying global semantic information, such as spatio-temporal relationships, enhancing the understanding of skeleton data on a broader scale.  The DA-Flow architecture, depicted in Fig.~\ref{block}, integrates these elements by employing residual links to maintain both local and global insights throughout the model. This strategy captures nuanced details and overarching patterns in the skeletal data. In the inference process, we evaluate each pose segment individually. 

In the following sections, we offer a succinct overview of normalizing flow and GCN, then introduce the DAM. The DAM's effectiveness is highlighted through the synergistic functions of \textbf{\textit{Skeleton Attention}} and \textbf{\textit{Frame Attention}}.

\subsection{Affine Transformation-Based Normalizing Flow}
\begin{figure}[!t]
   \centering
   \includegraphics[scale=0.3]{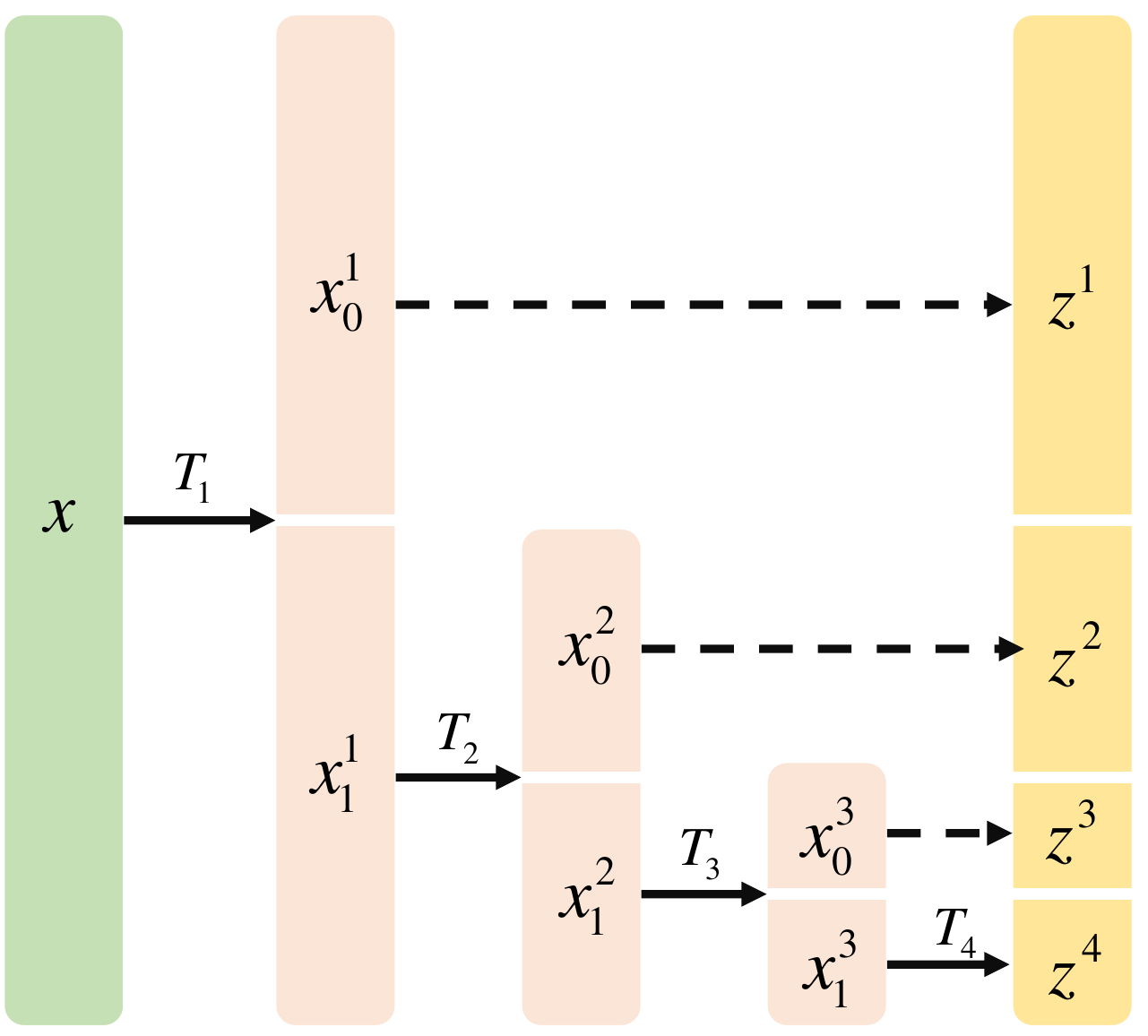}
    \caption{ Multiscale architecture with four levels as introduced in \cite{29NVP}. First, the entire input $x$ is transformed by $T_1$. The result is then split up into two parts of which one of them is factored out immediately and the other one is further processed by $T_2$. This process is repeated a few times until the desired depth is
reached. The input is drawn in green, intermediate results are in red, and the components of the final variable $z$ are yellow. } 
    \label{nf}

\end{figure}
Normalizing flow provides a robust framework for modeling complex probability distributions by transforming a base distribution through a sequence of invertible and differentiable operations. Consider a random variable $Z \in \mathbb{R}^D$ with a probability density function $p_Z(z)$. Introducing a bijective map $T: \mathbb{R}^D \leftrightarrow \mathbb{R}^D$ parameterized by a neural network, denoted as $x = {T}(z)$, the transformed random variable $X$ has a tractable density derived through the Jacobian determinant, expressed as:

\begin{equation}
    p_X(x) = p_Z(z) \left|\det J_{T}(z)\right|^{-1},
\end{equation}
where $J_T(z)$ denotes the Jacobian matrix of $T$ at $z$. The log-probability of the normalizing flow is given by:

\begin{equation}
    \log p_X(x) = \log p_Z(z) - \log \left|\det J_{T}(z)\right|.
    \label{lognf}
\end{equation}

To enhance computational efficiency, transformations \( T \) with triangular Jacobian matrices are preferred. These matrices simplify the calculation of the log-determinant to just the sum of diagonal elements. Mathematically, given an input vector $x$ split into two parts, $x_0$ and $x_1$, where $x_1$ undergoes transformation and $x_0$ remains unchanged, the transformation of the $i$-th level can be expressed as:

\begin{align}
y_1^i &= s^i(x_0^{i-1}) \otimes x_1^{i-1} + t^i(x_0^{i-1}) \\
y_2^i &= x_0^{i-1},
\end{align}
where $\otimes$ stands for Hadamard product, $y_1^i$ and $y_0^i$ represent the transformed parts of the input, and $s^i(x_0^{i-1})$ and $t^i(x_0^{i-1})$ are scale and translation functions which typically implemented as neural networks taking $x_0$ as input. This process is repeated a few times until the desired depth is reached. A diagram of an affine transformation-based multiscale normalizing flow is shown in Fig.~\ref{nf}.

\subsection{Graph Convolutional Networks}
 In the context of processing skeletal data with GCNs \cite{35gcn}, the graph convolution operation can be efficiently represented in matrix form. This representation is crucial for capturing the simultaneous aggregation of node features across the entire graph, leveraging the structural properties encoded in the adjacency matrix. The operation is mathematically expressed as:

\begin{equation}
Y =\hat{D}^{-1/2} \hat{A} \hat{D}^{-1/2} X W,
\end{equation}
where $\hat{A} = A + I$ represents the adjacency matrix $A$ augmented with the identity matrix $I$, incorporating self-connections to include each node's features in the update. The matrix $\hat{D}$ is the degree matrix corresponding to $\hat{A}$.  We use the distance partitioning strategy in \cite{yan2018stgcn} to construct the adjacency matrix. The $X\in\mathbb{R}^{C \times V}$ is a skeleton diagram, where $C$ and $V$ represent the number of channels and skeletal joints respectively, with each column representing the features of a node. $W$ is the weight matrix, containing the trainable parameters that transform node features from one layer to the next. $Y$ denotes the output of GCN.

\subsection{Dual Attention Module}

\subsubsection{Rethinking GCN and TCN}
GCN is primarily utilized for processing data in the spatial dimension which applies convolutional operations on graph-structured data to extract features, where the graph structure represents complex spatial relations, such as the connections within human skeleton nodes. TCN focuses on extracting features in the temporal dimension. By applying $1 \times 1$ convolution operations on time-series data, TCN traverses the temporal dimension to capture features that change over time. 

This dual approach facilitates the extraction of localized nodal information through graph convolution and the distillation of temporal dynamics on an individual joint basis via $1 \times 1$ temporal convolution. However, we believe that GCN and TCN have the following two main shortcomings in processing spatial-temporal skeleton diagrams for SVAD:

\begin{enumerate}[1.]
    \item \textit{Restricted Receptive Field:} GCN's ability to capture interactions between distinct body parts such as the head and hand, is limited by its reliance on heuristic partitioning. This leads to insufficient connectivity between such parts when they are not grouped within the same partition. Furthermore, after GCN processing, TCN's $1 \times 1$ convolution operates within a constrained time domain receptive field. Consequently, both GCN and TCN exhibit notably limited post-convolution receptive fields, which limit their effectiveness in capturing spatio-temporal co-occurrence relationships.
\end{enumerate}
\begin{enumerate}[2.]
    \item \textit{Inadequate cross-dimension interaction capture:}  
    Both GCN and TCN fall short of effectively capturing the interaction across dimensions in skeleton data, including the relationships between channels and skeletal joints, as well as between channels and frames. This limitation is especially evident in applications like fall detection, where variations along the \textit{$y$}-axis among different skeletal nodes and over time are more pronounced than those along the \textit{$x$}-axis. It is essential to capture the relationship across dimensions to better comprehend the intricate spatial-temporal patterns present in such data
    \end{enumerate}

 Consequently, these two shortcomings together lead to a failure to capture global high-level semantic information, which is crucial for understanding complex relational dynamics, such as the changing positional relationship between the head and hand across multiple frames. Although the receptive field can be increased by designing multi-scale architectures, the additional computational and storage overhead is unacceptable for VAD which requires timely intervention. 
 
 Addressing this challenge requires an effective approach that not only prioritizes the extraction of local information through GCN but also incorporates a robust mechanism for capturing global semantic insights with minimal computational and storage overhead. 

To this end, we introduce the DAM, a sophisticated post-processing attention mechanism for GCNs. This module is meticulously engineered to fortify the model's ability to assimilate and process global information after GCN's local relation extraction, ensuring a comprehensive data understanding. The DAM features two components: \textbf{\textit{Skeleton Attention}} and \textbf{\textit{Frame Attention}}. They operate on an input tensor $X \in \mathbb{R}^{C \times T \times V}$, where $C$, $T$, and $V$ represent the number of channels, frames, and skeletal joints, respectively. \textbf{\textit{Skeleton Attention}} orchestrates the generation of a 2-D attention map $M_{s} \in \mathbb{R}^{1 \times C \times V}$, whereas \textbf{\textit{Frame Attention}} yields an attention map $M_{t} \in \mathbb{R}^{1 \times T \times C}$. The output of DAM is obtained by initially conducting broadcast multiplication of each attention map with its corresponding input tensor. Subsequently, the results from \textbf{\textit{Skeleton Attention}} are concatenated with those from \textbf{\textit{Frame Attention}}. Illustrative details of the computational process for each branch are depicted in Fig.~\ref{block}. Comprehensive elaboration on each module will be provided in the ensuing sections.

\begin{figure}[t]
   \centering
   \includegraphics[scale=0.22]{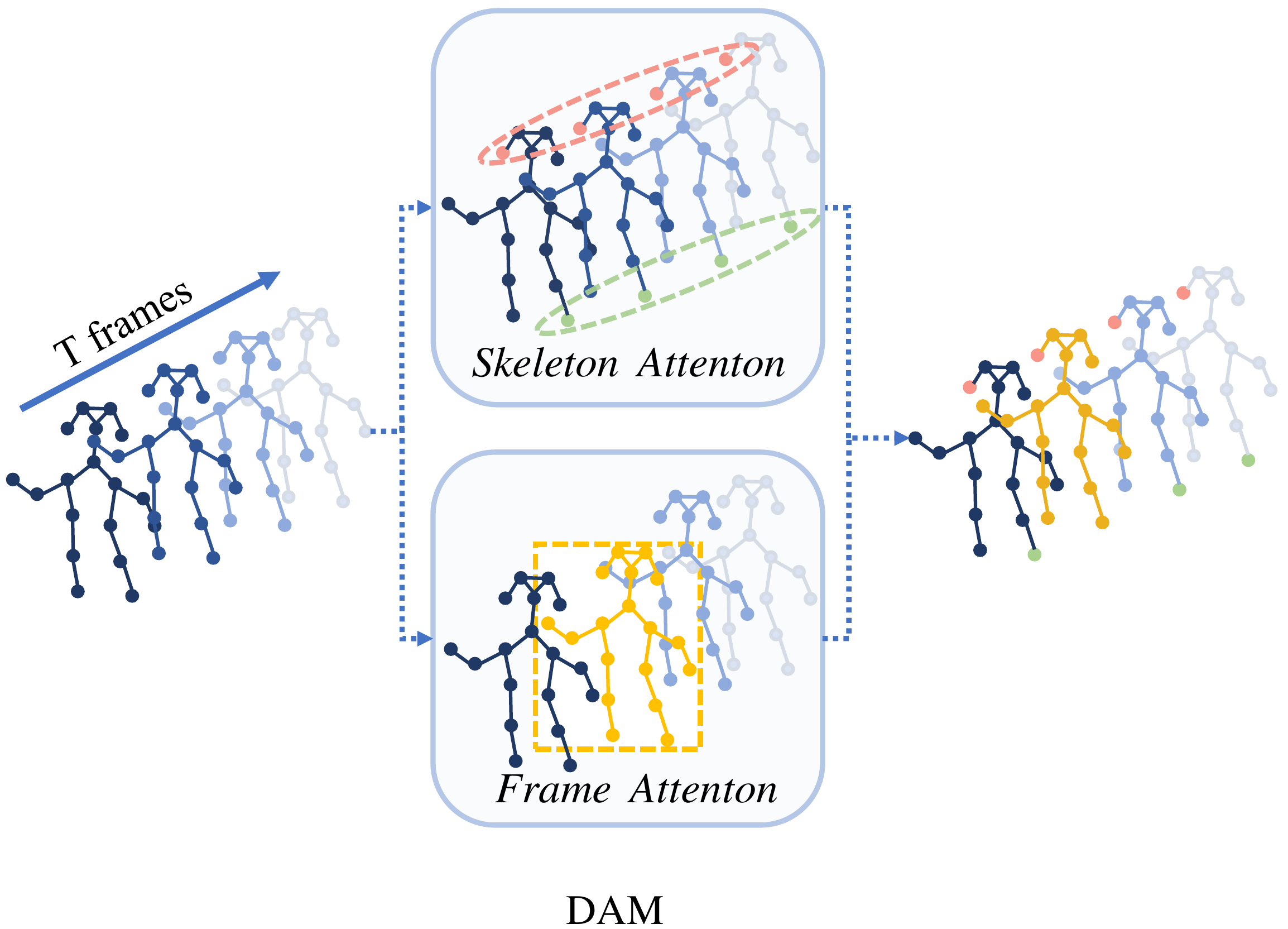}

    \caption{The figure for DAM: DAM captures the global information across all input frames and cross-GCN-partition: 
    The red and green boxes represent the most noteworthy skeleton nodes called \textbf{\textit{Skeleton Attention}} and the orange box denotes the frame that is most noteworthy for determining the anomaly called \textbf{\textit{Frame Attention}}. } 
    \label{dual}

\end{figure}
\subsubsection{Skeleton Attention}

\textbf{\textit{Skeleton Attention}} is designed to capture broader relationships across fixed partitions and channel-skeletal joints interaction within skeleton data, focusing on identifying the most critical joints following GCN processing, thereby addressing the question of \textit{\textbf{`Which skeletal joints are more crucial?'}}

The initial step within the \textbf{\textit{Skeleton Attention}} branch involves permuting the input tensor $X$ to $X_{s} \in \mathbb{R}^{T \times C \times V}$. This permutation effectively rotates $X$ by \(90^\circ\) anti-clockwise along the $V$ axis. Subsequently, max pooling operations are applied along the $T$ axis to extract the `sharpest' features across all input frames. The resulting pooled features are concatenated and fed through a 2-D convolutional layer, generating a 2-D spatial attention map $M_{s} \in \mathbb{R}^{1 \times C \times V}$.

The output of the \textbf{\textit{Skeleton Attention}} branch is obtained by broadcasting the element-wise multiplication of $X_{s}$ and $M_{s}$ along the $T$ axis:

\begin{equation}
\overline{Y_s} = \overline{\sigma(f(p(X_{s}))) \odot X_{s}} = \overline{M_s \odot X_s}, 
\label{sa}
\end{equation}
where the symbol $\odot$ denotes broadcast element-wise multiplication. The function \(p\) is defined as:

\begin{equation}
p(X) = \textit{MaxPool}(X),
\end{equation}
where \(\textit{MaxPool}(X)\) represents max pooling across channels. When applied to $X_{s} \in \mathbb{R}^{T \times C \times V}$, \(p(X_s)\) results in a  $ 1 \times C \times V $ tensor. The function \(f\) corresponds to a convolution operation with a \(3 \times 7\) filter, and the symbol \(\sigma\) denotes the sigmoid function.

The tensor $\overline{Y_{s}}$ in equation~\eqref{sa} is obtained by rotating the tensor \(Y_{s}\) \(90^\circ\) clockwise along the \(V\) axis to retain the original input shape of $C \times T \times V$.

As illustrated in Fig.~\ref{dual}, the green and red boxes produced by the upper branch represent the output of \textbf{\textit{Skeleton Attention}}.

\subsubsection{Frame Attention}

\textbf{\textit{Frame Attention}} decides to explore `\textbf{\textit{Which frames are more crucial}}' and channels-frames interaction relationships. This branch aims to generate an attention map $M_{t} \in \mathbb{R}^{1 \times T \times C}$ in refining the learning process for \textbf{\textit{Frame Attention}}.
The structure of \textbf{\textit{Frame Attention}} is identical to the structure of \textbf{\textit{Skeleton Attention}}.
The output of the \textbf{\textit{Frame Attention}} branch is obtained by :
\begin{equation}
\begin{aligned}
\overline{Y_t} = &  \overline{\sigma(f(p(X_{t}))) \odot X_{t}} )
=&\overline{M_{t_1} \odot X_{t_1}}, 
\label{fa}
\end{aligned}
\end{equation}
where \(X_{t} \in \mathbb{R}^{V \times T \times C}\) represents the input tensor \(X\) rotated \(90^{\circ}\) anti-clockwise along the \(T\) axis.  $\overline{Y_{t}}$ in equation ~\eqref{fa} represents tensor \(Y_{t}\) rotated \(90^{\circ}\) clockwise along the \(T\) axis to retain the original input shape of $C \times T \times V$.
In Fig.~\ref{dual}, the orange box produced by the lower branch represents the output of \textbf{\textit{Skeleton Attention}}.

Overall, the final output \(Y\) of the DAM can be expressed as:
\begin{equation}
Y =  \frac{1}{2}  \left( \overline{Y_{s}} + \overline{Y_{t}} \right) + X .
\label{da}
\end{equation}
Here, \(X\) is the residual output after GCN processing, which is used to preserve the local features obtained by GCN.

\subsubsection{Computational Complexity and Parameter Number Analysis}

To the best of our knowledge, DAM represents the pioneering effort to introduce an attention mechanism specifically designed for lightweight SVAD solutions. This initiative is compared against the parameter counts of prevalent convolution-based attention mechanisms extensively utilized in the field of computer vision. These include the CBAM \cite{woo2018cbam}, Triplet Attention \cite{17}, Channel Attention \cite{hu2018senet}, Coordinate Attention \cite{hou2021coordinate} for an input tensor of shape \([1,2,24,18]\). Here, the configuration denotes a batch size of 1, a channel dimension of 2 (capturing \(x, y\) coordinates), a temporal dimension consisting of 24 frames, and 18 skeletal nodes.

\begin{figure}[t]
    \centering
        \includegraphics[scale=0.35]{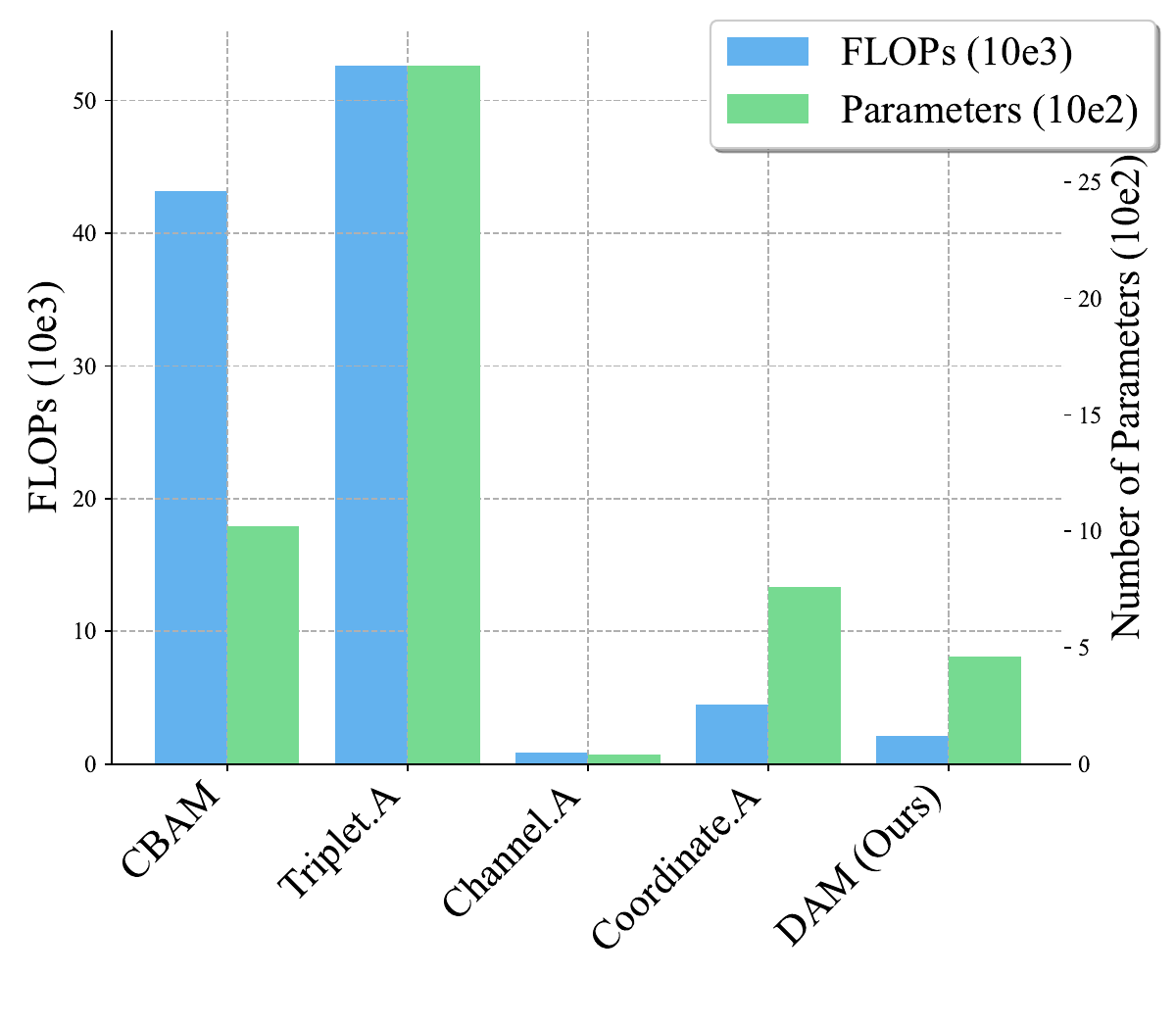}
\vspace{-3mm}
        \caption{Comparison of Module FLOPs and Parameters with Popular Attention Mechanisms: CBAM \cite{woo2018cbam}, Triplet Attention \cite{18}, SE Net \cite{hu2018senet}, Coordinate Attention and \cite{hou2021coordinate} on Skeleton Data with Input Shapes of $[1,2,24,18]$.}
        \label{flops}
\vspace{-3mm}
\end{figure}

It is observed in Fig.~\ref{flops} that DAM surpasses only the channel attention mechanism \cite{hu2018senet} in terms of computational complexity and parameter count. But we emphasize that for the skeleton data, regarding the post-GCN module, channel attention appears overly simplistic for only 2 channels, entirely neglecting temporal information. It is equivalent to a `convolution of channels'.  In contrast, the parameter counts and floating point operations (FLOPs) for alternative attention mechanisms are substantially greater. This makes them less suitable for deployment in scenarios demanding real-time anomaly detection. In subsequent experiments, we will verify the advantages of DAM for processing spatio-temporal skeletal data for VAD over other kinds of attention used in computer vision.

\section{experiment}

In the experimental section, we begin by delineating the dataset and evaluation metrics utilized, followed by a detailed description of the experimental setup. Subsequently, we conduct a comparative analysis of the state-of-the-art methods in VAD, encompassing both RGB image and video stream-based approaches as well as those leveraging posture and skeleton data. Further, we assess the efficacy of the DAM relative to other conv-based attention mechanisms within our framework. The robustness of our model is then evaluated against challenges such as random noise and the presence of negative samples. Finally, we undertake a series of ablation studies to investigate the impact of various parameters, including the number of coupling layers, different coupling layers, the influence of sliding window size, convolution kernel size, and different pooling methods on the experimental outcomes. Finally, we opted not to train our model but instead conducted random projections to assess the significance of the statistical characteristics of normal skeleton data for VAD.

\begin{table}[!b]
\fontsize{8}{13.8}\selectfont
\centering

 \renewcommand{\arraystretch}{0.7}
\setlength{\tabcolsep}{3pt}
\centering
\caption{Overview of five datasets used in the experiment. (Frames)}
\begin{tabular}{@{}lcccccc@{}}
\toprule
{}     & \textbf{Training} & \textbf{Validation} & \textbf{Test}  & \textbf{Normal} & \textbf{Abnormal}\\ \midrule
\textbf{STC}    & 274,515 & -          & 42,883 & 300,308 & 17,090    \\
\textbf{UBnormal }  & 116,087 & 28,175   & 92,640 & 147,887 & 89,015    \\ 
\textbf{HR-STC}    & 274,515 & -          & 38,697 & 297,000 & 16,122    \\
\textbf{HR-UBnormal}  & 116,087 & 28,175 & 90,489 & 147,887 & 86,864    \\
\textbf{UCSD Ped2}  & 2550 & - & 2010 & 2924 & 1636    \\
\bottomrule
\label{over}
\end{tabular}

\end{table}
\begin{table*}[t]
	\centering
 \renewcommand{\arraystretch}{1.2}
\setlength{\tabcolsep}{8pt}
	\caption{Comparison of DA-Flow against other video anomaly detection methods on five datasets: ShanghaiTech (STC), HR-ShanghaiTech (HR-STC), UBnormal, HR-UBnormal, and UCSD Ped2. The left side represents the type of training data used: RGB images and video streams, human pose. The best results are in bold and the second-best results are underlined.
 }
\begin{tabular}{cc|cc|ccccc}
\hline
\textbf{Method} &\textbf{Venue} & \textbf{RGB} & \textbf{Pose} & \textbf{STC} & \textbf{HR-STC} & \textbf{UBnormal} & \textbf{HR-UBnormal}& \textbf{Ped2}\\
\hline
Conv AE \cite{36convae} & \textit{CVPR'16} & \checkmark  &- &70.4 & 69.8 & - &- &90.0\\
Pred \cite{37pred} & \textit{CVPR'18} & \checkmark  &- &72.7 & 72.8 & - &- &95.4\\
MPED-RNN \cite{mpedrnn} & \textit{CVPR'19} & -  &\checkmark & 73.4& 75.5 & 60.6 & 61.2 &- \\
GEPC \cite{gepc} & \textit{CVPR'20} & -  &\checkmark & 75.2& 74.8 & 53.4 & 55.2  &-\\
MT Prediction \cite{multi} & \textit{WACV'20} & -  &\checkmark & 76.3& 77.4 & - &-\\
Multispace VAD \cite{zhang2020multispace} & \textit{TCSVT'20} & \checkmark  &- & 73.6& - & -& - &95.2\\
Normal Graph \cite{normalgraph} & \textit{Neurocomputing’21} & -  &\checkmark & 74.1& 76.5 & - & - &-\\
HF$^2$-VAD \cite{hfvad} & \textit{ICCV’21} & - & \checkmark &-&76.2& - & -  & \textbf{99.3} \\
PoseCVAE \cite{posecvae} & \textit{ICPR’21} & -  &\checkmark & 74.9& 75.1 & - & - & - \\
SSMTL \cite{ssmtl} & \textit{CVPR’21} & \checkmark  &- &82.4& - & 55.4 & - & 92.4 \\
HSGCNN \cite{zeng2021hierarchical} & \textit{TCSVT'21}& -  &\checkmark & 81.8  &83.4 & - &-  &97.7\\
BiPOCO \cite{bipoco} & \textit{Arxiv’22} & -  &\checkmark & 73.7& 74.9 & 50.7 & 52.3 & - \\
STGCAE-LSTM \cite{stgcaelstm} & \textit{Neurocomputing ’22} & -  &\checkmark & 75.6& 77.2 & - & - & - \\
Jigsaw \cite{jigsaw}& \textit{ECCV’22}  &\checkmark  &- & 84.3& -& 55.6 & - & \underline{98.8} \\
SSMTL++ \cite{SSMTL++}& \textit{CVIU’23} & \checkmark  &-&83.8& - &  62.1 & -  & - \\
MoPRL \cite{moprl}& \textit{TCSVT’23} &-  &\checkmark&83.4& 84.3 &  - & - & - \\
MGGAN-CL \cite{li2023multigan} & \textit{TCSVT’23} & \checkmark  &- & 73.6& - & - & - & 96.5 \\
COSKAD \cite{COSKAD} & \textit{Arxiv’23} & -  &\checkmark &-& 77.1 & 65.0 & 65.5 & - \\
STG-NF \cite{stgnf} & \textit{ICCV’23} & - &\checkmark &85.9& \underline{87.4} & 71.8 & - & - \\
MoCoDAD \cite{mocodad} & \textit{ICCV’23} & -  &\checkmark &-& 77.6 & 68.3 & 68.4& -  \\
TrajREC  \cite{stergiou2024traj} & \textit{WACV’24} & -  &\checkmark &-& 77.9 & 68.0 & 68.2 & - \\
VADiffusion \cite{liu2024vadiffusion} & \textit{TCSVT’24} & \checkmark   &-&71.7& - & - & - & 98.2 \\
 GiCiSAD \cite{karami2024gici} & \textit{Arxiv’24} & -   &\checkmark&-& 78.0 & 68.6 & 68.8 & - \\
Two Stream \cite{cao2024twostream} & \textit{TIP’24} & \checkmark   &-&83.7& - & - & - & 97.1 \\
 \rowcolor{gray!15}
\textbf{FA-Flow}(Ours) &  &-  &\checkmark &\underline{86.1}& 87.3 & \underline{73.7} & \underline{73.7} & 94.1\\
 \rowcolor{gray!15}
\textbf{SA-Flow}(Ours) &  &-  &\checkmark &82.2& 82.4 &71.0 & 70.6 & 92.3\\
 \rowcolor{gray!30}
\textbf{DA-Flow}(Ours) &  &-  &\checkmark &\textbf{86.5}& \textbf{87.8} & \textbf{74.1} & \textbf{74.2} & 95.3\\
\hline
\label{all}
\end{tabular}
\end{table*}

\subsection{Dataset and Metrics}

Our experiments utilize several publicly available datasets for video anomaly detection, namely, ShanghaiTech Campus\cite{20}, UBnormal\cite{21} and UCSD Ped2 \cite{li2013ped2}, We use the filtering method proposed by \cite{25} to filter the ShanghaiTech Campus dataset and the UBnormal dataset. This division yielded two specialized subsets: HR-STC and HR-UBnormal, which exclusively encompass anomalies induced by human actions, thereby segregating them from the broader spectrum of anomalies found within the original compilations of the datasets. An overview of the dataset is presented in Table~\ref{over}.

Notably, our training process exclusively utilizes normal samples, with anomalous events introduced solely during the testing phase.

\textbf{ShanghaiTech Campus} \cite{20}: Encompasses 330 training and 107 test videos across 13 scenes, each video at $856 \times 480$ resolution. Known for its dynamic scenes with complex lighting and varied camera angles, this dataset includes diverse anomalies such as car invasions and robberies.

\textbf{UBnormal} \cite{21}: A synthetic and open-set benchmark containing 268 training, 64 validation, and 211 test videos. It uniquely offers annotations at both frame and pixel levels and covers a wide array of normal and abnormal behaviors.

\textbf{UCSD Ped2} \cite{li2013ped2}: A component of the UCSD anomaly detection dataset which focuses on pedestrian areas with 16 training and 12 test videos at $360 \times 240$ resolution. This dataset targets U-VAD and features regular pedestrian activities alongside anomalies such as vehicles in pedestrian zones with frame-level annotations.

\begin{figure}[!h]
    \centering
    \includegraphics[width=0.5\textwidth]{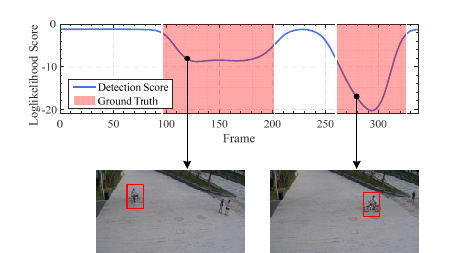}\hfill
    \includegraphics[width=0.5\textwidth]{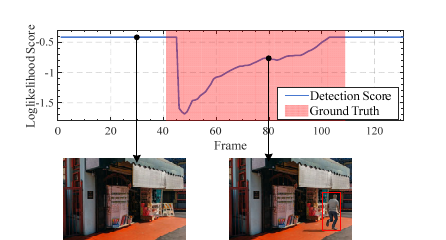}\hfill
    \includegraphics[width=0.5\textwidth]{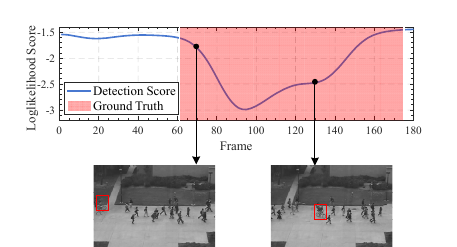}
    \caption{Log-likelihood score examples in STC, UBnormal, and UCSD ped2 datasets: This figure illustrates our method's ability to identify anomalies in video frames accurately. Frames are scored based on the minimal pose score, with ground truth anomalous frames highlighted in red, demonstrating effective anomaly flagging.} 
    \label{fig: example1}
\end{figure}

The model's performance is evaluated using the most common metric called the microscopic receiver operating characteristic Area Under the Curve (AUC) metric.
Here, we report the log-likelihood scores obtained on STC, UBnormal, and UCSD Ped2 datasets. An illustrative example is depicted in Fig.~\ref{fig: example1}. Our model effectively detects the anomaly in both space and time.

\subsection{Implementation Details}
For frame-wise skeleton detection, we utilize AlphaPose \cite{34alphapose} in combination with YOLOX \cite{32yolox} for tracking skeletons in video sequences. Our model adopts a prior distribution of $\mathcal{N}(3, I)$, utilizes $K=8$ flow steps, and integrates a single GCN block followed by the DAM at each step. The temporal segment length $T$ is set to $24$ for the STC and HR-STC datasets, $T=4$ for the UBnormal and HR-UBnormal datasets, and $T=20$ for the UCSD Ped2 dataset.

Experiments were conducted using an Adam optimizer with a learning rate of $5 \times 10^{-4}$, across 8 epochs. Batch size is set to 256. The computational setup included an 11th Gen Intel Core\texttrademark{} i5-11400F CPU operating at 2.60GHz, complemented by an Nvidia GeForce RTX 3080 GPU.

\subsection{Comparison With State-Of-The-Art Methods}

For an exhaustive comparison, we selected a diverse range of baseline models. This includes models based on RGB images and video streams, such as Conv AE \cite{36convae}, Pred \cite{37pred}, Multispace VAD \cite{zhang2020multispace}, SSMTL \cite{ssmtl}, Jigsaw \cite{jigsaw}, SSMTL++ \cite{SSMTL++}, MGGAN-CL \cite{li2023multigan}, VADiffusion \cite{liu2024vadiffusion} and Two Stream \cite{cao2024twostream}. Additionally, we have included models that emphasize human pose data for anomaly detection, including MPED-RNN \cite{mpedrnn}, GEPC \cite{gepc}, MT Prediction \cite{multi}, PoseCVAE \cite{posecvae}, Normal Graph \cite{normalgraph}, HF$^2$-VAD \cite{hfvad}, HSGCNN \cite{zeng2021hierarchical}, BiPOCO \cite{bipoco}, STGCAE-LSTM \cite{stgcaelstm}, MoPRL \cite{moprl}, COSKAD \cite{COSKAD}, STG-NF \cite{stgnf}, MoCoDAD \cite{mocodad}, GiCiSAD \cite{karami2024gici} and TrajREC  \cite{stergiou2024traj}. The performance metrics presented in Table \ref{all} are sourced either directly from the original papers or from studies that compare the methodologies of these papers.

Our DA-Flow model achieves the highest AUC scores, with 86.5\% on the STC dataset and 87.8\% on the HR-STC dataset. On the UBnormal and HR-UBnormal datasets, DA-Flow scores 74.1\% and 74.2\%. It surpasses the SOTA method in RGB images on the STC dataset by 2.1\% \cite{jigsaw}. On the UBnormal dataset, DA-Flow surpasses the SOTA method \cite{SSMTL++} in RGB images by 11.2\%. Additionally, our model outperforms the leading pose-based method \cite{stgnf} by 0.6\% on the STC dataset and 2.3\% on the UBnormal dataset. 

Within the UCSD Ped2 dataset, our method exhibits performance inferior to methods based on RGB images. This disparity can be attributed primarily to the dataset's modest resolution ($360 \times 240$ pixels), which considerably hampers the efficacy of skeleton detection algorithms, thereby amplifying the margin of error.  Despite these challenges, our method still manages to achieve commendable performance levels. Notably, our method has the minimal training parameter count among the evaluated skeleton-based methodologies as shown in Table \ref{ped}. 


\begin{table}[t]
\centering
\vspace{-5mm}
\renewcommand{\arraystretch}{1.2}
\setlength{\tabcolsep}{3pt}
\caption{Comparison of Model Training Parameters with SOTA SVAD methods: HSGCNN \cite{zeng2021hierarchical}, STG-NF \cite{stgnf}, MoCoDAD \cite{mocodad}, HSGCNN \cite{zeng2021hierarchical}, COSKAD \cite{COSKAD}, TrajREC \cite{stergiou2024traj}, and GiCiSAD \cite{karami2024gici}. }

\begin{tabular}{lcccc}
\toprule
\textbf{Method} & \textbf{Venue} & \textbf{Training Param (K)} \\
\midrule
 HSGCNN \cite{zeng2021hierarchical} & \textit{TCSVT'21}        & 2.331          \\ 
STG-NF \cite{stgnf} & \textit{ICCV'23}        & 0.616          \\ 
  MoCoDAD \cite{mocodad} & \textit{ICCV'23}        & 142.2          \\ 
   COSKAD \cite{COSKAD} & \textit{Arxiv'23}        & 240.1          \\ 
      TrajREC \cite{stergiou2024traj} & \textit{WACV'24}        & 4900          \\ 
      GiCiSAD \cite{karami2024gici} & \textit{Arxiv'24}        & 82.6         \\
\rowcolor{gray!30}
\textbf{DA-Flow (ours)}  &           &  \textbf{0.488}              \\
\bottomrule
\end{tabular}
\vspace{-5mm}
\label{ped}
\end{table}

In our study, we also experimented with two variations of our model, named Skeleton-Attention Flow (SA-Flow) and Frame-Attention Flow (FA-Flow), which isolate components \textbf{\textit{Skeleton Attention}} and \textbf{\textit{Frame Attention}}, respectively. Our findings reveal that FA-Flow, which emphasizes time domain information and inter-channel relationships, outperforms SA-Flow. This suggests that temporal dynamics and the interaction between different data channels are crucial for accurately detecting anomalies in video sequences.

\subsection{Compare DAM with Convolutional Attention Mechanism }
In this section, we evaluate the performance of our model's DAM against established convolution-based attention mechanisms: CBAM \cite{woo2018cbam}, Triplet Attention \cite{18}, SE Net \cite{hu2018senet}, and Coordinate Attention \cite{hou2021coordinate}. These methods are integrated with GCN on three distinct datasets to compare their effectiveness.
\begin{table}[htbp]
\centering

 \renewcommand{\arraystretch}{1.2}
\setlength{\tabcolsep}{5pt} 
\caption{Comparison of Popular Conv-based Attention Mechanisms: CBAM \cite{woo2018cbam}, Triplet Attention \cite{18}, SE Net \cite{hu2018senet}, Coordinate Attention and \cite{hou2021coordinate} with DAM on Their Performance on STC, UBnormal, and UCSD Ped2 Datasets. The best results are in bold, and the second-best results are underlined.}
\label{table:att}

\begin{tabular}{lccc}
\toprule
\textbf{Method} & \textbf{STC} & \textbf{UBnormal} & \textbf{Ped 2}\\
\midrule
GCN & 81.3 & 70.2 & 92.1 \\
GCN+CBAM & 85.7 & 72.3 & 93.4 \\
GCN+Triplet Attention & \underline{86.3} & \underline{74.0} & \underline{94.4} \\
GCN+Channel Attention & 81.1 & 70.2 & 92.2 \\
GCN+Coordinate Attention & 85.3 & 73.3 & 93.2 \\
\rowcolor{gray!30}
\textbf{GCN+Dual Attention (ours)} & \textbf{86.5} & \textbf{74.1} & \textbf{95.3} \\
\bottomrule
\end{tabular}

\end{table}

As shown in Table~\ref{table:att}, our observations indicate that our proposed method outperforms existing techniques. Notably, Triplet Attention \cite{18} is the closest to our results, at the cost of the latter's significant computational and storage demands (as depicted in Fig.~\ref{flops}). This also confirms the importance of capturing cross-dimension interaction relationships.
\subsection{Robustness Analysis}

Next, we assess the robustness of our model, which showed superior performance on four datasets: STC, HR-STC, UBnormal, and HR-UBnormal. We introduced random Gaussian noise into the skeleton data to simulate inaccuracies from pose estimators and incorporated anomalous data into the training set. This approach tested the model's resilience under varied conditions.

\subsubsection{Gaussian Noise}
\begin{figure}[h]
	\centering

	\includegraphics[scale=0.45]{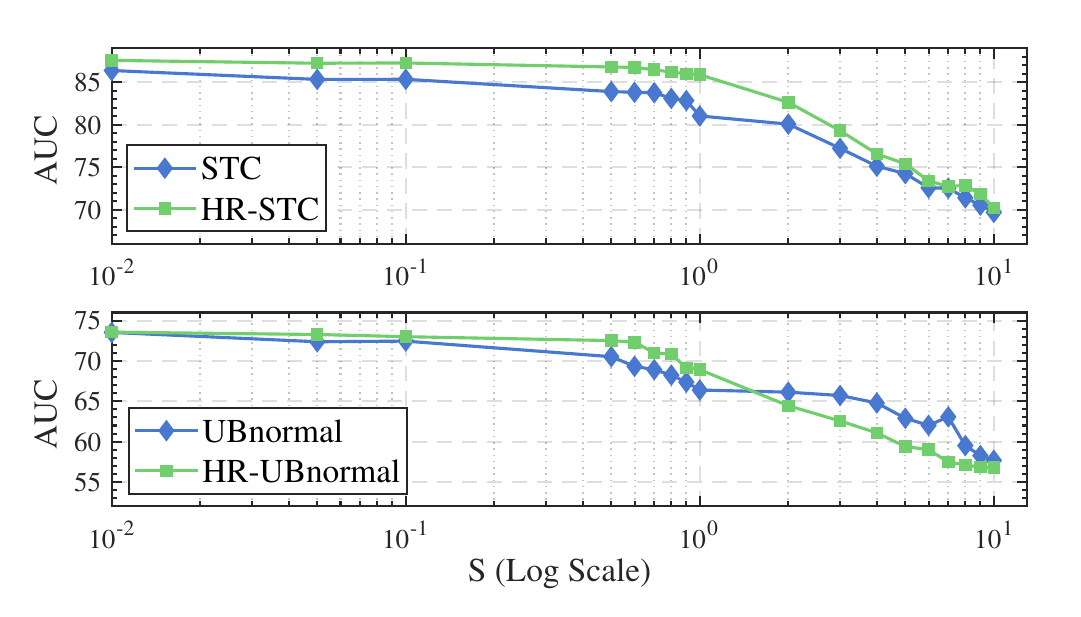}

	\caption{Effect of the various scales of Gaussian noise on the DA-Flow detection performance. Our model demonstrates substantial robustness against a considerable amount of noise.}
	\label{fig：Gaussian noise}
\end{figure}

To evaluate the impact of errors from pose estimators, we simulated these errors by introducing Gaussian noise to each key point. Specifically, we applied varying scales of noise, \(S \cdot u\), where \(u\) follows a normal distribution, \(\mathcal{N}(0, I)\). As depicted in Fig.~\ref{fig：Gaussian noise}, for the STC and UBnormal datasets, the AUC decreased by less than 3\% for a noise scale below 1. Conversely, for HR-STC and HR-UBnormal, the decrease was around 6\%. This outcome underscores our model's resilience to significant key point noise, indicating it can still perform effectively even with less accurate pose estimators. The AUC drops considerably at higher noise levels. Nonetheless, across all datasets, our model demonstrated remarkable robustness, maintaining high performance in VAD. 
\subsubsection{Anomalous Data}
\begin{figure}[!h]
	\centering
	\includegraphics[scale=0.45]{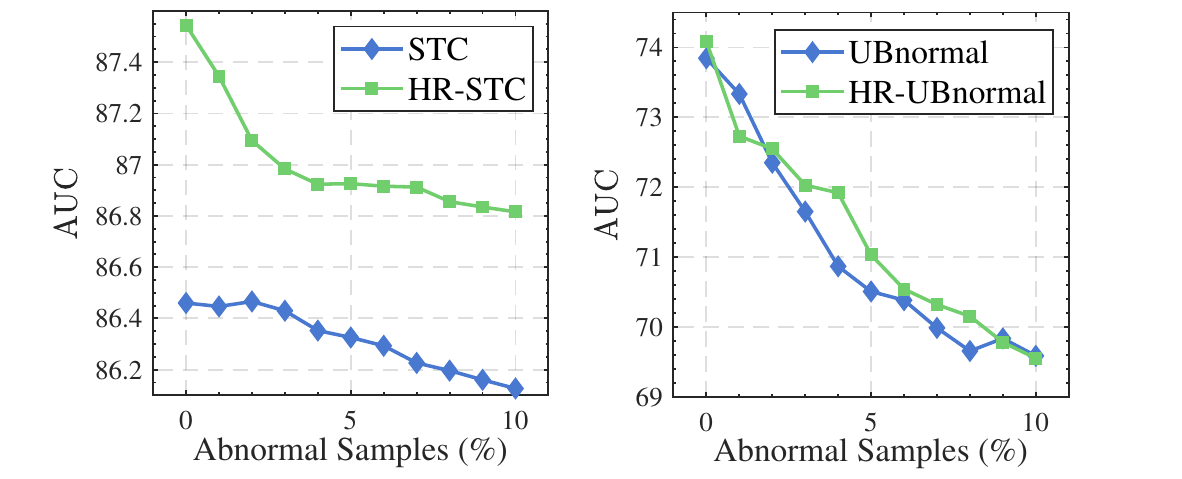}

	\caption{Effect of the proportion of training data added to abnormal data on the DA-Flow detection performance. Our model maintains robustness even when a percentage of the samples were mislabeled as abnormal.}
	\label{fig：Abnormal Samples}

\end{figure}

To further evaluate the robustness of our model, we conducted tests by incorporating varying proportions of anomalous data into the training set to simulate a contaminated training data set, as illustrated in Fig.~\ref{fig：Abnormal Samples}. When applied to the STC dataset, this procedure resulted in a modest performance decline of approximately 1\%. Similarly, on the HR-STC dataset, we observed a decrease of 0.7\%. In contrast, when the same proportion of anomalous data was introduced into the UBnormal and HR-UBnormal datasets, the performance deterioration was more pronounced, around 4\%. This difference could be due to UBnormal's synthetic nature and our model's reliance on skeleton pose data. In this case, our model also shows strong robustness

\subsection{More Ablation Experiment}
In this section, we provide further experiments used to evaluate different model components :
\subsubsection{Number of Coupling Layers. }
\begin{figure}[!t]
	\centering
	\includegraphics[scale=0.45]{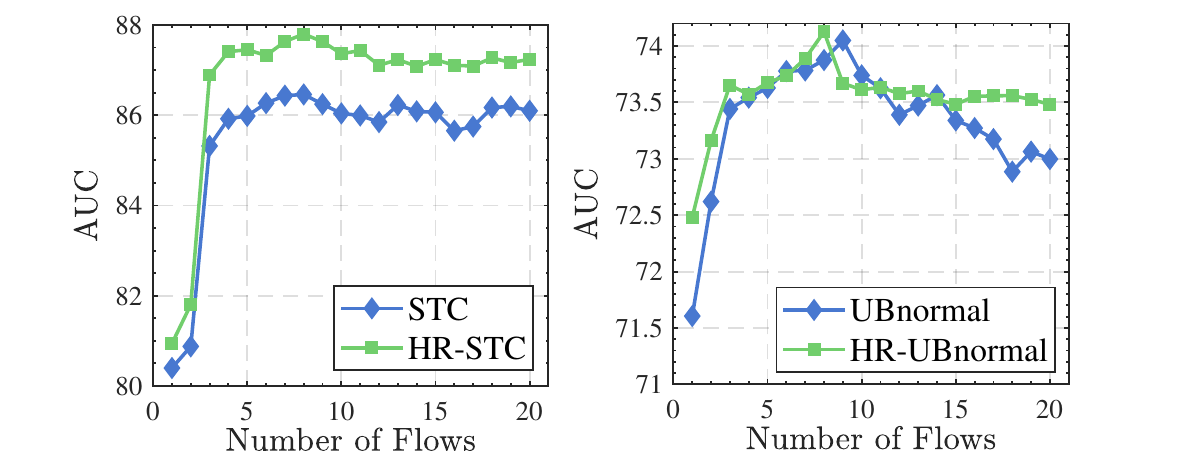}
	\caption{Effect of the number of affine coupling layers on the model detection performance. The peak value is reached at $K=8$.}
	\label{fig: Number of flows}
\end{figure}

In our experimental analyses conducted on four distinct datasets, we first explored the impact of varying the number of affine coupling layers ($K$) on our models' efficacy.  Fig.~\ref{fig: Number of flows} demonstrates that model performance significantly improves with incremental addition of flow layers, reaching an optimum at $K=8$. Beyond $K=8$, we observed a plateau in performance metrics which implies that the incorporation of additional affine coupling layers beyond $K=8$ does not yield substantial enhancements. Our findings underscore that while increasing the number of affine coupling layers augments the flexibility of the normalizing flow's transformation, enabling more variables to be aligned with the prior distribution, this advantage caps at $K=8$. Beyond this point, the transformation process becomes overly rigid, detrimentally affecting the model's capacity for generalization.

\subsubsection{Different Coupling Layers}
\begin{figure}[!t]
	\centering
	\includegraphics[scale=0.45]{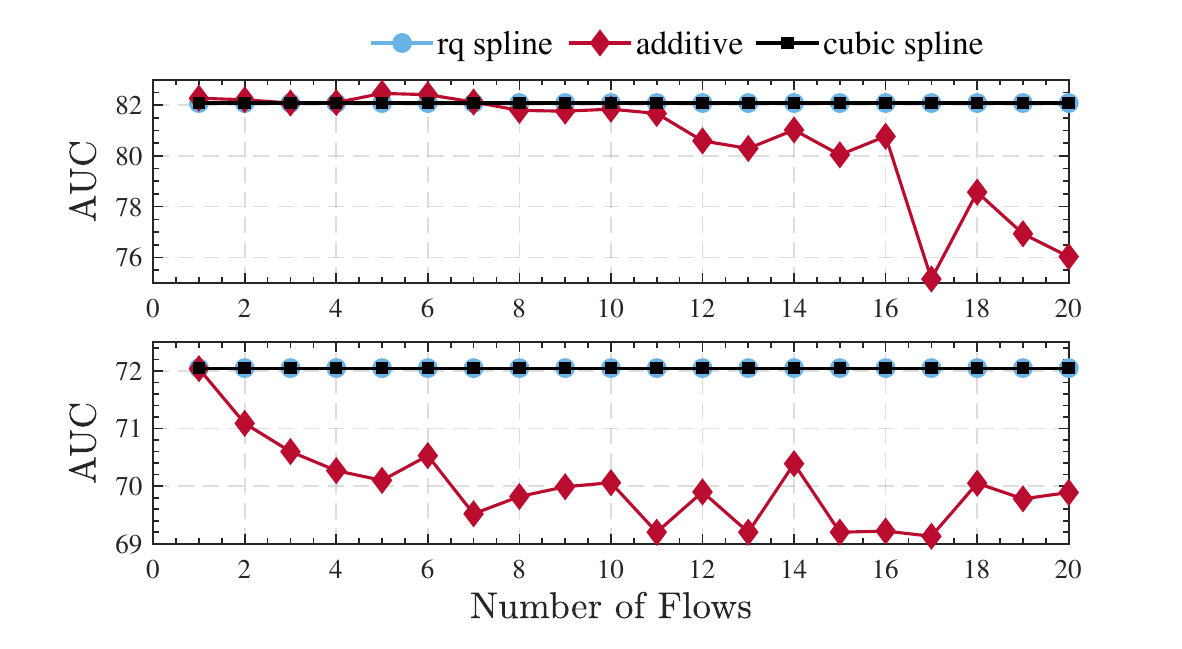}
	\caption{The impact of varying the number of coupling layers in DA-Flow: More flexible transformations bolster the model's robustness at the potential expense of its generalization capabilities.}
	\label{fig:coupling layer}
\end{figure}
In our study, we also carried out comparative experiments to assess the effects of various coupling layers on the final detection outcomes. Utilizing the Glow \cite{27glow} architecture, we evaluated the performance implications of three distinct types of coupling layers: the additive coupling layer introduced by NICE \cite{28nice}, the rational-quadratic (RQ) neural spline \cite{30nsf} coupling layer and the cubic spline coupling layer \cite{durkan2019cubic} across the aforementioned datasets. Generally, transformations implemented using cubic spline coupling layers and rational-quadratic neural spline coupling layers exhibit greater flexibility compared to those employed by additive and affine coupling layers.

As shown in Fig.~\ref{fig:coupling layer}, our findings reveal that the enhanced flexibility afforded by the rational-quadratic and cubic neural spline coupling layers leads to a tighter alignment of samples with the prior distribution. This close alignment mitigates the impact of varying the number of transformation modules on detection performance, thereby enhancing the model's robustness. However, this increase in robustness might come at the cost of diminished generalization capabilities. This observation further substantiates our earlier conclusion drawn from the comparative analysis of different quantities of affine coupling layers: overly precise mapping can detrimentally affect the model's performance.


\subsubsection{Sliding Window Steps}

\begin{figure}[!h]
	\centering
	\includegraphics[scale=0.45]{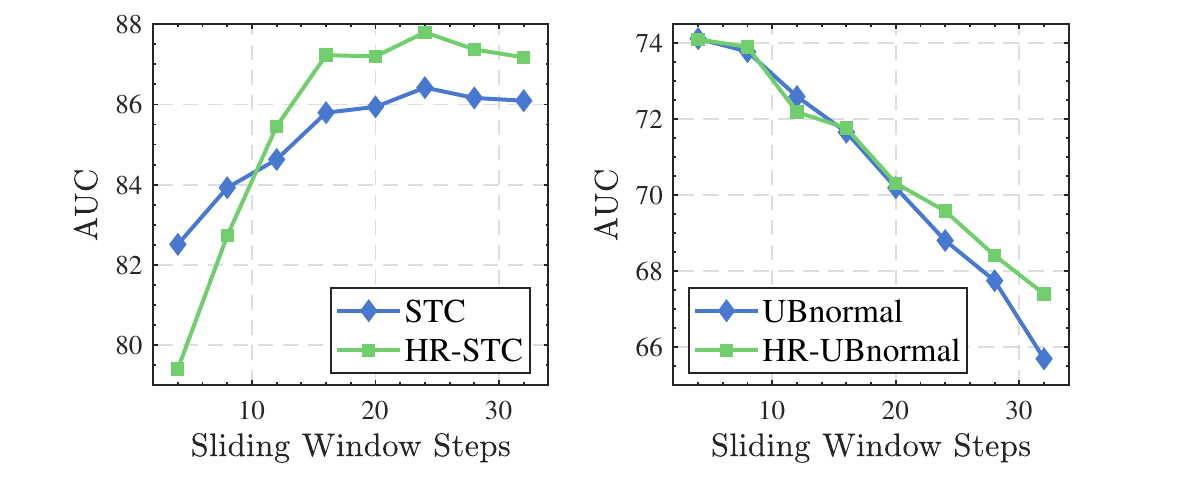}
	\caption{Effect of the number of sliding window step sizes on the model detection performance. On the STC and HR-STC datasets, optimal performance is achieved at a step size of $T=24$. For the UBnormal and HR-UBnormal datasets, the peak performance occurs at $
 T=4$.}

	\label{fig:Sliding Window Steps}
\end{figure}

We also investigated how changes in the sliding window step size affect our model's efficacy. This step size, which represents the quantity of skeleton data frames gathered over a set time span, plays a pivotal role in the model's ability to detect patterns effectively. As depicted in Fig.~\ref{fig:Sliding Window Steps}, on the STC and HR-STC datasets, an increase in the sliding window step size correlates with improved model performance, reaching a peak at $T=24$. In contrast, for the UBnormal and HR-UBnormal datasets, we observe an enhancement in performance with an increment in step size up to $T=4$, beyond which, a further increase leads to a decline in performance. 


\subsubsection{Convolution Kernel Size of DAM. }

In this experiment, we examine the impact of convolution kernel sizes in the DAM on its performance across the STC, UBnormal, and UCSD Ped2 datasets. We particularly focus on kernel size selection in the channel dimension, noting that a kernel size of 1 leads to ignoring the correlation between the $x$ and $y$ coordinates of skeleton data. This approach parallels that adopted by the TCN. Our empirical findings indicate a decline in performance metrics, with the AUC metric experiencing a decrease of approximately 1.1\% for the STC dataset, 1.4\% for the UBnormal dataset, and 2.1\% for the UCSD Ped2 dataset. These results support our hypothesis that the interplay between the $x$ and $y$ coordinates is pivotal.

Additionally, we investigated the effects of varying the convolution kernel size along another dimension, specifically testing kernel sizes of 3, 5, 7, and 9. Our analysis demonstrates that a kernel size of 7 yields the best performance, a finding consistent with those reported in previous studies such as CBAM \cite{woo2018cbam} and Triplet Attention \cite{17}. Performance improves progressively from a kernel size of 3 to 7, followed by a performance decrease at a kernel size of 9.
\begin{table}[h]

\centering
 \renewcommand{\arraystretch}{1.2}
\setlength{\tabcolsep}{8pt} 
\caption{Comparison of Different Kernel Sizes: $1\times7$, $3\times3$, $3\times5$, $3\times7$, and $3\times9$ in DAM on Their Performance on STC, UBnormal, and UCSD Ped2 Datasets. The best results are in bold, and the second-best results are underlined.}
\label{table:kernel-size-performance}

\begin{tabular}{lccc}
\toprule
\textbf{Kernel Size} & \textbf{STC} & \textbf{UBnormal} & \textbf{Ped 2}\\
\midrule
$1\times7$ & 85.4 & 72.7 & 93.2 \\
$3\times3$ & 85.5 & 72.3 & 93.1 \\
$3\times5$ & 85.9 & 73.5 & 94.4 \\
\rowcolor{gray!30}
\textbf{$3\times7$ (used)} & \textbf{86.5} & \textbf{74.1} & \textbf{95.3} \\
$3\times9$ & \underline{86.1} & \underline{73.7} & \underline{94.7} \\
\bottomrule
\end{tabular}
\end{table}
\subsubsection{Pooling Type of DAM}

In this experiment, we explore the impact of maximum and average pooling on the performance of DAM. Generally, maximum pooling retains the most distinct or `sharp' attributes of the input data, ensuring that the strongest features are emphasized. On the other hand, average pooling computes the mean of all features within a specific dimension, thereby smoothing out the data representation. 

We conducted a series of experiments to evaluate the impact of maximum pooling, average pooling, and a combination of both pooling methods on the performance of DA-Flow applied to the STC, UBnormal, and UCSD Ped2 datasets.

\begin{table}[ht]
    \centering
    \fontsize{8}{10}\selectfont
 \renewcommand{\arraystretch}{1.2}
    \setlength{\tabcolsep}{3pt}
    \caption{Comparison of Pooling Types in DAM on Model Parameters, FLOPs, and Performance on STC, UBnormal, and Ped 2 datasets. The best results are highlighted in bold, and the second-best results are underlined.}
    \label{pool}
    \begin{tabular}{c|ccc|c|c}
        \toprule
        \multirow{2}{*}{\textbf{Pooling type}} & \multicolumn{3}{c|}{\textbf{AUC}} & \multirow{2}{*}{\textbf{Flops (K)}}&\multirow{2}{*}{  \textbf{Param (K)}} \\
        \cline{2-4}
        & \textbf{STC} & \textbf{UBnormal} & \textbf{Ped 2} & \\
        \midrule
        AvgPool & 86.1 & 73.7 & 94.2 & 2.10 & 0.46   \\
        \rowcolor{gray!20}
        \textbf{MaxPool (used)} & \textbf{86.5}& \underline{74.1} & \textbf{95.3} & \textbf{2.10} & \textbf{0.46}  \\
        Both & \textbf{86.5} & \textbf{74.3} & \underline{95.1} & \underline{3.86} & \underline{0.88}  \\
        \bottomrule
    \end{tabular}
\end{table}
As shown in Table~\ref{pool}, the performance using average pooling in DAM is not as good as the performance using maximum pooling. In particular, when both types of pooling are used, performance on STC datasets is comparable to maximum pooling alone. The UBnormal data set is slightly higher than using maximum pooling, while the UCSD Ped2 data set is not as good as maximum pooling. However, the cost is nearly double the FLOPS overhead and the number of parameters. 

\subsubsection{Zero training}
To assess the significance of the statistical characteristics of normal skeleton data in VAD, we adopt a unique approach. We do not train our model. Instead, we test it immediately after a random setup. We conduct 100 experiments on the above five datasets and calculate the average results. Table~\ref{tab: zero-training} presents these results. On the UBnormal dataset, our method without training achieves an AUC of 69.97\%, outperforming SSMTL++v1 \cite{SSMTL++} by about 7\% and diffusion-based SVAD methods MoCoDAD \cite{mocodad} and TrajREC \cite{stergiou2024traj} 2\%. On the STC dataset, our method attains an AUC of 78.24\%, which is superior to most traditional methods that utilize images and video streams. These findings emphasize the effectiveness of using skeleton data in distinguishing anomalies. We posit that the skeleton diagram is mapped into an alternate space via the random projection feature of normalizing flow without reducing dimensionality. This transformation results in distinct statistical properties and superior-level sets between normal behavior and abnormal behavior, which are particularly effective in distinguishing positive anomalous behaviors. Based on this observation, we believe that the focus on SVAD in future work should not be solely on reconstruction and prediction, or a combination, but rather focusing on the statistical characteristics of normal samples.

\begin{table}[h]
\setlength{\tabcolsep}{4pt}
    \centering
    \caption{AUC Performance Comparison across Datasets with Zero-Training. }
    \label{tab: zero-training}
    \begin{tabular}{@{}lccccc@{}}
        \toprule
        & \textbf{STC} & \textbf{HR-STC} & \textbf{Ubnormal} & \textbf{HR-Ubnormal} & \textbf{UCSD-Ped2 }\\
        \midrule
        \textbf{AUC (\%)} & 78.24 & 79.54 & 69.97 & 70.07 & 84.96 \\
        \bottomrule
    \end{tabular}
\end{table}

\section{Conclusion}
This paper introduces the Dual  Attention Module (DAM), a lightweight and effective component designed to capture the key global features and cross-dimension interaction relationships in skeleton data across frames for VAD. Furthermore, we present DA-Flow, a novel lightweight model that integrates DAM post-GCN within a normalizing flow framework. Our method significantly improves global correlation and effectively captures key spatio-temporal co-occurrence relationships through \textbf{\textit{Frame Attention}} and \textbf{\textit{Skeleton Attention}}, resulting in enhanced anomaly detection capacity. 
Extensive evaluations of four datasets confirm the superior performance and robustness of DA-Flow with minimal parameters. Moreover, We discover that even without training, simply employing random projection on skeleton data enables us to achieve substantial anomaly detection capabilities.


\bibliographystyle{IEEEtran}
\bibliography{main}

\end{document}